# Expert Opinion Elicitation for Assisting Deep Learning based Lyme Disease Classifier with Patient Data

Sk Imran Hossain[a], Jocelyn de Goër de Herve[b,c], David Abrial[b,c], Richard Emillion[d], Isabelle Lebert[b,c], Yann Frendo[a,b], Delphine Martineau[e], Olivier Lesens[f,g], Engelbert Mephu Nguifo[a,*]

[a]Université Clermont Auvergne, CNRS, ENSMSE, LIMOS, F-63000 Clermont-Ferrand, France
[b]Université Clermont Auvergne, INRAE, VetAgro Sup, UMR EPIA, 63122 Saint-Genès-Champanelle, France
[c]Université de Lyon, INRAE, VetAgro Sup, UMR EPIA, F-69280 Marcy l'Etoile, France
[d]Université d'Orléans, 45067 Orléans, France
[e]Infectious and Tropical Diseases Department, CHU Clermont-Ferrand, Clermont-Ferrand, France
[f]Infectious and Tropical Diseases Department, CRIOA, CHU Clermont-Ferrand, Clermont-Ferrand, France
[g]UMR CNRS 6023, Laboratoire Microorganismes: Génome Environnement (LMGE), Université Clermont Auvergne, Clermont-Ferrand, France
*Corresponding author
phone: +33473407629, fax: +33473407639, email: engelbert.mephu_nguifo@uca.fr

## Abstract

Diagnosing erythema migrans (EM) skin lesion, the most common early symptom of Lyme disease using deep learning techniques can be effective to prevent long-term complications. Existing works on deep learning based EM recognition only utilizes lesion image due to the lack of a dataset of Lyme disease related images with associated patient data. Physicians rely on patient information about the background of the skin lesion to confirm their diagnosis. In order to assist the deep learning model with a probability score calculated from patient data, this study elicited opinion from fifteen doctors. For the elicitation process, a questionnaire with questions and possible answers related to EM was prepared. Doctors provided relative weights to different answers to the questions. We converted doctors' evaluations to probability scores using Gaussian mixture based density estimation. For elicited probability model validation, we exploited formal concept analysis and decision tree. The elicited probability scores can be utilized to make image based deep learning Lyme disease pre-scanners robust.

## 1. Introduction

Lyme disease is one of the most common tick-borne diseases that infects a large no of people each year in Europe and America [1]. Lyme disease is caused by pathogenic bacteria of the *Borrelia burgdorferi* sensu lato group and mostly manifests itself with erythema migrans (EM) skin lesions in the early stage [2,3]. The EM usually goes away after a few months or weeks, but the infection from Lyme disease spreads to harm either the nerve system, joints, heart, eyes, or skin[4]. In the early stage of Lyme disease, antibiotics can be an effective therapy option. Most North American and European standards employ a two-tier serology test to identify antibodies against *Borrelia burgdorferi* sensu lato in order to diagnose Lyme disease [5,6]. However, a serology test is recommended only in the absence of EM, because early serology can lead to false negatives and has low sensitivity (40-60%) [5]. *Borrelia burgdorferi* sensu lato can also be detected directly via microscopy, culture, or PCR [6]. Bacterial culture the gold standard of microbiological diagnosis requires special media and laboratory expertise [5]. In practice, light microscopy based detection is not feasible [6]. PCR-based diagnosis is equally problematic, with a wide range of sensitivity [6]. Because of the limitations of direct detection technologies, physicians are not always able to employ them. Therefore, early detection of EM is critical for avoiding the long-term consequences of Lyme disease.

Existing works on early Lyme disease prediction using artificial intelligence techniques only utilize images of EM skin lesions whereas doctors believe corresponding patient data should also be considered to strengthen the predictive performance [7,8]. Training a multimodal deep learning model utilizing both images and patient data requires a dataset of lesion images with associated patient data. Even though EM image datasets are available, creating a dataset with patient data linked with each lesion image would take much time.

Expert opinion elicitation can be effective when high quality data is difficult to collect [9]. Expert opinion elicitation and aggregation processes can be classified into two categories: behavioral and mathematical approaches [10]. The behavioral approach tries to produce group consensus among experts whereas, the mathematical approach combines subjective probabilities from experts using mathematical methods (some form of averaging). Expert elicitation proved effective for medical diagnosis and decision making [9,11,12]. In this study, we elicited opinions from fifteen expert dermatologists to assist the image-based EM classifier with additional patient data. The traditional expert elicitation process of collecting probability estimates for all possible cases is time consuming and difficult for doctors. Therefore, we opted for a more relaxed approach of relative weight assignment to different answers to the questions and converted the doctor's evolutions to EM probabilities utilizing Gaussian mixture based density estimation (described in Section 2.1). To validate the elicited probability model and explain its behavior to the experts we utilized formal concept analysis (described in Section 2.3) and decision tree (described in Section 2.4).

The rest of the paper is structured as follows: Section 2 provides the required theoretical background; Section 3 describes the expert elicitations process; Section 4 contains a discussion on the results obtained; finally, Section 5 provides concluding remarks.

## 2. Background

The required theoretical concepts to understand the rest of the paper are briefly described in the following subsections.

### 2.1. Gaussian Mixture Model

A Gaussian mixture model (GMM) is a probability density function represented as the weighted sum of component Gaussian densities [13]. The mixture represents a normally distributed overall population whereas the components represent subpopulations within the whole population. For one-dimensional data, a GMM with $M$ components can be defined as:

$$\hat{f}_{GMM}(x) = \sum_{m=1}^{M} \emptyset_m \, \mathcal{N}(x \mid \mu_m, \sigma_m) \qquad (1)$$

where, $\emptyset_m$ is the mixture weight of $m$-th component $\kappa_m$ satisfying $\sum_{m=1}^{M} \emptyset_m = 1$ and $\mathcal{N}(x \mid \mu_m, \sigma_m)$ is the distribution of a Gaussian component with mean $\mu_m$ and standard deviation $\sigma_m$ defined as:

$$\mathcal{N}(x \mid \mu_m, \sigma_m) = \frac{1}{\sigma_m \sqrt{2\pi}} e^{-\frac{1}{2}\left(\frac{x - \mu_m}{\sigma_m}\right)^2} \quad (2)$$

Expectation-Maximization, an iterative unsupervised learning technique can be used to determine the parameters of GMM [14]. Steps involved in Expectation-Maximization for $n$ data points $X = \{x_t \mid t = 1, \dots, n\}$ are:

- Guess initial values for $GMM$ parameters denoted by $\hat{\mu}_m, \hat{\sigma}_m,$ and $\widehat{\emptyset}_m$ respectively.
- Expectation step: calculate $\hat{\gamma}_{t,m}$, the probability of a point $x_t$ being generated by $\kappa_m$

$$\hat{\gamma}_{t,m} = \frac{\widehat{\emptyset}_m \, \mathcal{N}(x_t \mid \hat{\mu}_m, \hat{\sigma}_m)}{\sum_{r=1}^{M} \widehat{\emptyset}_r \, \mathcal{N}(x_t \mid \hat{\mu}_r, \hat{\sigma}_r)} \quad (3)$$

- Maximization step: Update GMM parameters using the following equations:

$$\hat{\mu}_m = \frac{\sum_{t=1}^{n} \hat{\gamma}_{t,m} x_t}{\sum_{t=1}^{n} \hat{\gamma}_{t,m}} \quad (4)$$

$$\hat{\sigma}_m = \sqrt{\frac{\sum_{t=1}^{n} \hat{\gamma}_{t,m} (x_t - \hat{\mu}_m)^2}{\sum_{t=1}^{n} \hat{\gamma}_{t,m}}} \quad (5)$$

$$\widehat{\emptyset}_m = \sum_{t=1}^{n} \frac{\hat{\gamma}_{t,m}}{n} \quad (6)$$

- Repeat Expectation and Maximization steps until the total likelihood $L$ converges, where

$$L = \prod_{t=1}^{n} \hat{f}_{GMM}(x_t) \quad (7)$$

Information criterion tests like Akaike Information Criteria (AIC) [15] and Bayesian Information Criteria (BIC) [16] can be used to select an appropriate GMM by penalizing the number of free parameters to prevent overfitting. AIC and BIC can be defined as:

$$AIC = 2p + 2 \ln L \quad (8)$$
$$BIC = p \ln L + 2 \ln L \quad (9)$$

where $p$ is the number of free parameters. The preferred GMM is the one with minimum AIC and BIC values.

## 2.2. Kernel Density Estimation

Kernel density estimation (KDE) is a non-parametric way of estimating the probability density function of an independent and identically distributed random variable [17,18]. For $n$ data points $X = \{x_t \mid t = 1, \dots, n\}$, KDE is calculated as:

$$\hat{f}_{KDE}(x) = \frac{1}{nh} \sum_{t=1}^{n} K\left(\frac{x - x_t}{h}\right) \quad (10)$$

where $h$ is the bandwidth and $K$ is the kernel function. If a Gaussian kernel function is used to estimate the density of univariate data then the bandwidth can be selected using Silverman's rule of thumb [19] as shown in the following equation:

$$h = 0.9 \min(\hat{\sigma}, \frac{IQR}{1,34}) n^{-1/5} \qquad (11)$$

where $IQR$ is the interquartile range and $\hat{\sigma}$ is the standard deviation of the samples.

## 2.3. Formal Concept Analysis and Concept Lattice

Formal concept analysis (FCA) is a method of generating a formal concept hierarchy from a set of objects and their properties [20]. Each concept represents objects that share a particular set of attributes. FCA computes concept lattice, a directed, acyclic graph by hierarchically ordering all formal concepts derived from tabular input data.

The notion of formal context is central to FCA. Formal context is a triple $\langle O, Y, I \rangle$ where $O$ is a set of objects, $Y$ is a set of attributes, and incidence $I \subseteq O \times Y$ is a binary relation. A pair $\langle A, B \rangle$ is a formal concept of $\langle O, Y, I \rangle$ provided that $A \subseteq O$, $B \subseteq Y$, $A^\uparrow = B$, and $B^\downarrow = A$ where

$A^\uparrow = \{y \in Y | for\ each\ o \in A: \langle o, y \rangle \in I\}$ and $B^\downarrow = \{o \in O | for\ each\ y \in B: \langle o, y \rangle \in I\}$.

Formal concepts are ordered naturally by subconcept-superconcept relation defined as follows:

$$\langle A_1, B_1 \rangle \leq \langle A_2, B_2 \rangle \iff A_1 \subseteq A_2 (\iff B_2 \subseteq B_1) \qquad (12)$$

For a formal context $\langle O, Y, I \rangle$ the set $\mathfrak{B}(O, Y, I)$ of all formal concepts with the ordering shown in Eq. (12) is called the concept lattice.

## 2.4. Decision Tree

Decision tree is an approach of representing a classifier as a recursive partition of instance space using a set of splitting rules [21]. These rules are easy to visualize and interpret with tree diagrams. Decision tree is a directed tree with no incoming edges at the root node and each of the other nodes has just one incoming edge. A decision or leaf or terminal node is a node without outgoing edges. All other nodes are called test or internal nodes. The instance space is divided into two or more sub-spaces by each test node based on a discrete function of input attribute values. Each decision node is given a class that corresponds to the best suitable target value. Instances are classified according to the test results by navigating from the tree's root to a leaf.

## 3. Elicitation Method

The details of our expert elicitation process like expert recruitment, questionnaire preparation, experts' opinion collection, and elicitation methods are described in the following subsections.

## 3.1. Expert Selection

The recruited experts are hospital practitioners who are infectious disease specialists or dermatologists working in reference centers for tick-borne diseases of France - Centres de Référence des Maladies Vectorielles liées aux Tiques (CRMVT) [22]. At a CRMVT steering committee meeting held in June 2021 with participants from all the reference centers, Professor Olivier Lesens (Infectious and Tropical Diseases Department, CRIOA, CHU Clermont-Ferrand, France) explained the importance of expert elicitation for calculating EM probability based on patient data and requested the interested experts to participate in the elicitation process. Fifteen experts agreed to participate. The experts did not receive any monetary benefits for participating in the elicitation process. Appendix Table A1 in the Supplementary Data file available at the link mentioned in the data statement lists the reference centers and the number of experts participating in the elicitation process.

### 3.2 Questionnaire and Experts' Evaluation

For the EM probability elicitation, a questionnaire was prepared based on questions about the context of onset and progression of the skin lesion that a physician usually asks when diagnosing EM. The questionnaire is based on a previous study concerning the collection of EM related data from rural areas of France [23]. The questionnaire was finalized through several meetings held in April 2020 among the doctors of CRMVT in Clermont-Ferrand and experts in tick ecology from the French national research institute for agriculture, food and the environment (INRAE) [24]. Experts who volunteered to participate in the elicitation process at the meeting in June 2021 agreed that there were many possible cases from the combination of the questions and answers, and it was time consuming and difficult for them to provide probability estimates for all those different cases. Therefore, experts agreed to independently assign relative weights to different possible answers associated with each question. The assigned weight values are in the range -1 to +3 (a higher value represents a higher contribution of the answer towards the possibility of the EM). The experts were contacted via email with detailed instructions to provide their weight attributions independently. Appendix Table B1 in the Supplementary Data file available at the link mentioned in the data statement lists the questions, answers, and weight attribution from the doctors.

After receiving the weight attributions from all the experts, they participated in a meeting in November 2021 and agreed that fever, fatigue, faintness, and headache should contribute equally if one or more of these answers were present and the contribution should be the average of these four answers. Therefore, the four answers were replaced with one, and the possible cases reduced to 1,536 from 12,288 cases. This modification is shown in Table 1.

### 3.3 Opinion Elicitation

Following are some notations used in the rest of the manuscript:

Set of doctors, $D = \{d_e | e = 1, \dots, 15\}$

Set of questions, $Q = \{q_i | i = 1, \dots, 6\}$

Set of possible cases, $C = \{c_l | l = 1, \dots, 1536\}$

Total number of answers corresponding to $q_i$ question $= n_{q_i}$

$j^{th}$ answer corresponding to $q_i$ question, $a_{j,q_i} = \begin{cases} 1, & \text{if the answer is true} \\ 0, & \text{otherwise} \end{cases}$

$$\text{where } j = 1, \dots, n_{q_i}$$

Weight assigned by doctor $d_e$ to $a_{j,q_i}$ answer $= w_{d_e, a_{j,q_i}}$

Table 1: Weight modified questionnaire and doctors' weight attribution for erythema migrans. The assigned weight values are in the range -1 to +3 (a higher value represents a higher contribution of the answer towards the possibility of the erythema migrans). $d_1$ to $d_{15}$ represents the doctors.

| Question | Answer | Weight Assigned by Doctors (Doctors' Evaluation) | | | | | | | | | | | | | | | |
|---|---|---|---|---|---|---|---|---|---|---|---|---|---|---|---|---|---|
| | | $d_1$ | $d_2$ | $d_3$ | $d_4$ | $d_5$ | $d_6$ | $d_7$ | $d_8$ | $d_9$ | $d_{10}$ | $d_{11}$ | $d_{12}$ | $d_{13}$ | $d_{14}$ | $d_{15}$ | Average |
| Other symptoms observed alongside the skin lesion ($q_1$) | No ($a_{1,q_1}$) | 0 | 0 | 3 | 0 | 0 | 1 | 2 | 1 | 2 | 1 | 2 | 1 | 1 | 2 | 3 | 1.27 |
| | Fever/ Fatigue/ Faintness/ Headache ($a_{2,q_1}$) | -0.25 | 0.25 | 0 | 0.75 | 0.75 | 0.25 | 0.75 | 0.75 | 0.75 | 1 | 1 | 0.25 | 1 | 0.25 | 0 | 0.5 |
| | Joint pain ($a_{3,q_1}$) | 1 | 1 | -1 | 2 | 2 | 1 | 1 | 1 | 1 | 1 | 1 | 1 | 1 | 0 | 0 | 0.87 |
| | Itching ($a_{4,q_1}$) | -1 | -1 | -1 | -1 | 1 | -1 | 0 | 0 | 0 | 1 | -0.5 | -1 | -1 | 1 | 0 | -0.3 |
| What was the maximum size of the red rash ($q_2$) | < 1 cm ($a_{1,q_2}$) | -1 | -1 | 0 | -1 | -1 | -1 | -1 | 0 | 0 | -1 | -1 | -1 | -1 | 1 | -1 | -0.67 |
| | 1 to 5 cm ($a_{2,q_2}$) | 1 | 1 | 1 | 0 | 1 | 0 | 1 | 1 | 2 | 1 | 1 | 1 | 1 | 2 | 1 | 1 |
| | > 5 cm ($a_{3,q_2}$) | 3 | 2 | 2 | 2 | 3 | 2 | 3 | 2 | 1 | 3 | 2 | 2 | 3 | 3 | 3 | 2.4 |
| | I do not know ($a_{4,q_2}$) | 0 | 0 | 0 | 0 | 0 | 0 | -1 | 1 | 0 | 0 | 0 | 0 | 0 | 0 | 0 | 0 |
| Is the size of the red rash increasing or has it gradually increased ($q_3$) | Yes ($a_{1,q_3}$) | 3 | 1 | 3 | 3 | 3 | 3 | 3 | 3 | 2 | 3 | 3 | 3 | 3 | 3 | 3 | 2.8 |
| | No ($a_{2,q_3}$) | 0 | -1 | -1 | -1 | -1 | -1 | -1 | 0 | -1 | 0 | -1 | -1 | -1 | 1 | -1 | -0.67 |
| | I do not know ($a_{3,q_3}$) | 0 | 0 | 0 | 0 | 0 | 0 | 0 | 1 | 0 | 0 | 0 | 0 | 0 | 0 | 0 | 0.07 |
| Have you seen a tick bite on this red rash in the past 30 days ($q_4$) | Yes ($a_{1,q_4}$) | 3 | 2 | 3 | 2 | 3 | 1 | 3 | 3 | 2 | 3 | 2 | 3 | 1 | 3 | 3 | 2.47 |
| | No ($a_{2,q_4}$) | 0 | 0 | 0 | 0 | 0 | 1 | 0 | 1 | 0 | 0 | -0.5 | -1 | 0 | 1 | 0 | 0.1 |
| Frequency of tick bites in the last 30 days before the appearance of the red rash ($q_5$) | Never ($a_{1,q_5}$) | -1 | -1 | 0 | 0 | -1 | 0 | 0 | 0 | 0 | 0 | -1 | -1 | 0 | -1 | 0 | -0.4 |
| | 1 time ($a_{2,q_5}$) | 0 | 0 | 2 | 1 | 1 | 1 | 1 | 1 | 2 | 1 | 1 | 1 | 1 | 2 | 1 | 1.07 |
| | 2 to 5 times ($a_{3,q_5}$) | 1 | 1 | 3 | 1 | 1 | 1 | 1 | 2 | 2 | 1 | 2 | 1 | 1 | 3 | 1 | 1.47 |
| | > 5 times ($a_{4,q_5}$) | 2 | 2 | 1 | 2 | 2 | 1 | 1 | 2 | 2 | 2 | 3 | 1 | 1 | 3 | 2 | 1.8 |
| Outdoor activities in the last 30 days before the onset of the red rash ($q_6$) | Yes ($a_{1,q_6}$) | 1 | 1 | 2 | 2 | 1 | 1 | 2 | 2 | 2 | 2 | 2 | 2 | 1 | 3 | 2 | 1.73 |
| | No ($a_{2,q_6}$) | -1 | -1 | -1 | -1 | -1 | 0 | -1 | 1 | -1 | -1 | -1 | -1 | 0 | -1 | 0 | -0.67 |

First, we summarized each of the 1,536 possible cases as a weight sum $s_{C_l}$ as shown in Eq. (13):

$$s_{C_l} = \sum_{i=1}^{|Q|} \sum_{j=1}^{n_{q_i}} a_{j,q_i} \times \left( \frac{1}{|D|} \sum_{d=1}^{|D|} w_{d_e, a_{j,q_i}} \right) \qquad (13)$$

The set of case weight sum is defined as $S = \{s_{C_l} | l = 1, \ldots, 1536\}$. Then, we normalized each case weight sum with min-max normalization as shown in the following equation:

$$\tilde{s}_{C_l} = \frac{s_{C_l} - min(S)}{max(S) - min(S)} \qquad (14)$$

The set of min-max normalized case weight sum is defined as $\tilde{S} = \{\tilde{s}_{C_l} | l = 1, \ldots, 1536\}$. We proposed three approaches to the experts to convert the normalized case weight sum to a probability score for EM. The following subsections explain the three approaches.

### 3.3.1 Cumulative Probability from Density Estimate Based on GMM

We modeled our normalized weight sum data density using a GMM with two components. The number of components was selected based on the intuition that there are two subpopulations within the data: one is the ill subpopulation and the other one is not ill subpopulation. The number of components was also supported by AIC and BIC values. Table 2 lists the selected parameters for the GMM.

Table 2: Parameters of Gaussian Mixture Model used to model the density of min-max normalized weight sum of erythema migrans cases. $\emptyset, \mu,$ and $\sigma$ represent mixture weight, mean and standard deviation respectively.

| Name | Components | $\emptyset_1$ | $\emptyset_2$ | $\mu_1$ | $\mu_2$ | $\sigma_1$ | $\sigma_2$ |
|---|---|---|---|---|---|---|---|
| Value | 2 | 0.364801 | 0.635199 | 0.359548 | 0.572878 | 0.128782 | 0.156241 |

The blue curve in Fig. 1 shows the estimated density function using GMM. We defined the cumulative probability [25] of a normalized case weight sum from the GMM density estimate as the probability of EM as shown in the following equation:

$$\hat{F}_{GMM}(x) = \int_{-\infty}^{x} \left( \sum_{m=1}^{2} \emptyset_m \, \mathcal{N}(x \mid \mu_m, \sigma_m) \right) dx \qquad (15)$$

### 3.3.2 Posterior Probability of a Case Belonging to the Ill Subpopulation of GMM

The first and second components of our GMM are shown in Fig. 1 with green and orange dotted lines respectively. If we assume that the second component represents the ill subpopulation then the posterior probability of a normalized case weight sum belonging to the second component [13] can be defined as the EM probability as shown in the following equation:

$$p(\kappa_2 \mid x) = \frac{\emptyset_2 \mathcal{N}(x \mid \mu_2, \sigma_2)}{\sum_{m=1}^{2} \emptyset_m \, \mathcal{N}(x \mid \mu_m, \sigma_m)} \qquad (16)$$

Non–Peer-reviewed Preprint Article

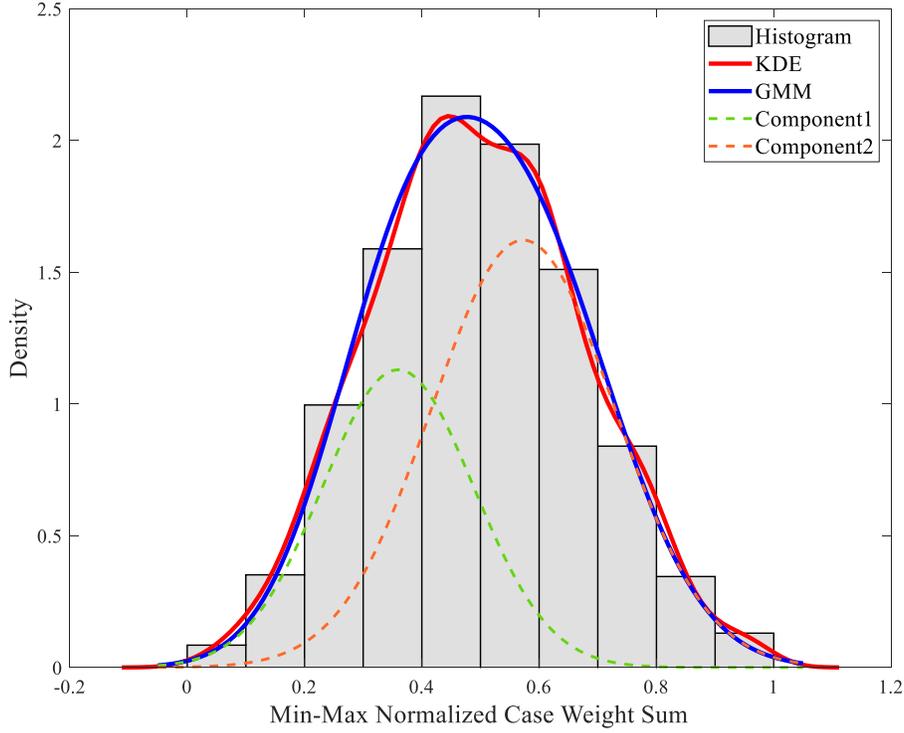

Fig. 1: Proposed approaches for expert opinion elicitation. GMM and KDE stand for Gaussian mixture model and kernel density estimation respectively.

### 3.3.3 Cumulative Probability from Density Estimate Based on KDE

We used a Gaussian kernel with bandwidth, h=0.03676 on our 1,536 data points for the probability density estimation of the normalized weight sum variable as shown in Eq. (4):

$$\hat{f}_{KDE}(x) = \frac{1}{1536 \times 0.03676} \sum_{l=1}^{1536} \frac{1}{2\pi} e^{-0.5\left(\frac{x-\tilde{s}_{C_l}}{0.03676}\right)^2} \quad (17)$$

The red curve in Fig. 1 shows the estimated density function. We defined the cumulative probability of a normalized case weight sum as the probability of having EM as shown in the following equation:

$$\hat{F}_{KDE}(x) = \int_{-\infty}^{x} \hat{f}_{KDE}(x) dx \quad (18)$$

### 4. Results and Discussion

We calculated EM probability score for all possible cases using the three approaches described in section 3.3.1, 3.3.2, and 3.3.3 and presented the results with explanations to the experts in a meeting held in May 2022. Fig. 2 shows the EM probability plot for all the cases using the three approaches. In the figure blue and red lines represent the probability scores based on density estimates from Gaussian mixture model (approach 1) and kernel density estimate (approach 2) respectively. The orange line represents probability scores based on the posterior probability of a case belonging to the second component i.e. the ill subpopulation of the Gaussian mixture model (approach 3). Results obtained from approach 1 and approach 2 are close because both of them are based on density estimates whereas, probability scores obtained from approach 3 are always higher than the other two approaches. Based on the results and explanations the experts came to a consensus on the use of approach 1 (described in section 3.3.1) mainly because the density estimate in approach 1 is smoother compared to approach 2 (described in section 3.3.2).

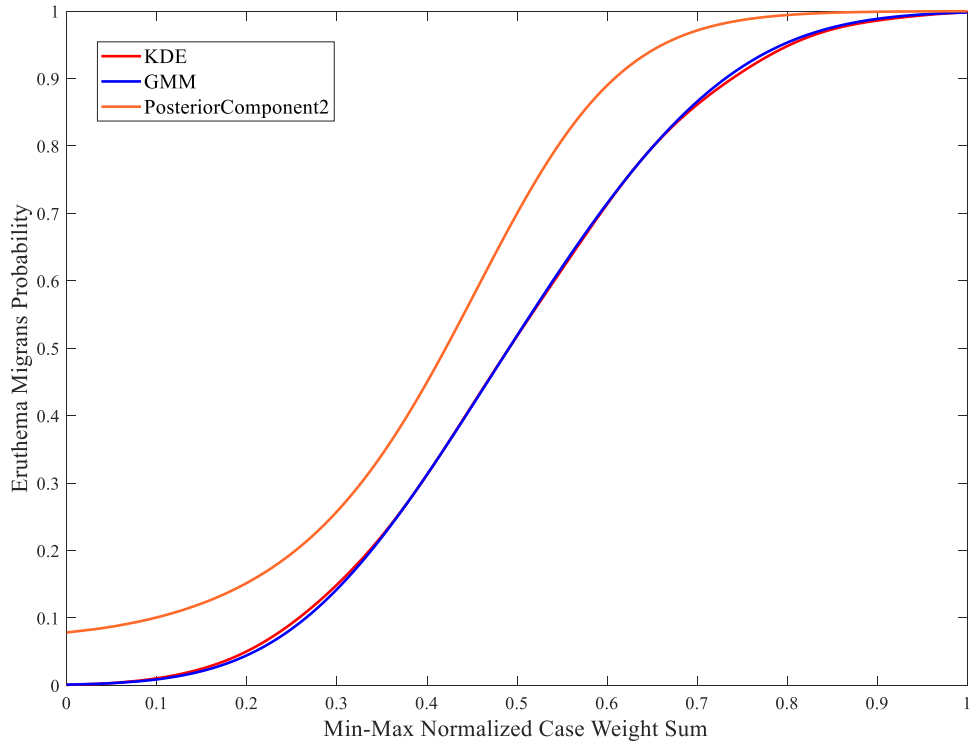

Fig. 2: Elicited erythema migrans probability plot. Blue and red lines represents the probability scores based on density estimates from Gaussian mixture model and kernel density estimate respectively. Orange line represents probability scores based on the posterior probability of a case belonging to the second component i.e. the ill subpopulation of the Gaussian mixture model.

To validate elicited model and explain its behavior to the experts first we used decision trees. For building the decision tree, we divided calculated EM probability scores into three categories: LOW (scores in the range [0, 0.33)), MEDIUM (scores in the range [0.33, 0.68)), and HIGH (scores in the range [0.68, 1]). Fig. 3 shows a pruned version of the decision tree for approach 1. In the figure, each node shows the majority category along with the percentage and number of cases belonging to each category. From the tree, we can see that the model assigns HIGH EM probability to cases whenever the first answer, "*yes*" to the third question, "*Is the size of the spot increasing or has it gradually increased*" is true. This behavior supports the doctors' opinion because the first answer to the third question has the highest weight given by most of the doctors.

To further explain the behavior of the model we utilized formal concept analysis (FCA) to find out questions and answers important for different probability groups. Fig. 4 shows a simplified FCA lattice view for the 162 cases belonging to the lowest probability score group in the range [0, 0.1) obtained from approach 1. In the figure, the top box of a node represents an attribute (answer) or a number of attributes, which are connected by lines, and the bottom box represents how many objects (cases) contain the corresponding attribute shown in the top box. In Fig. 4, we start with 162 cases in the root node. At the first level, the number inside the bottom box of a node represents how many cases out of 162 cases contain the corresponding answer shown in the top box. For example, the "*no*" answer to the question "*Outdoor activities in the last 30 days before the onset of the red spot*", $a_{2,q_6}$ is present in 145 cases. At the second level, each node represents how many cases contain two answers connected by a line. For example, $a_{2,q_4}$ and $a_{2,q_6}$ are jointly true in 128 cases. The rest of the FCA lattice is organized similarly. We can see from the figure that the answers common to most of these cases are the ones having lowest assigned weights or the opposites of the answers having highest assigned weights by most of the doctors.

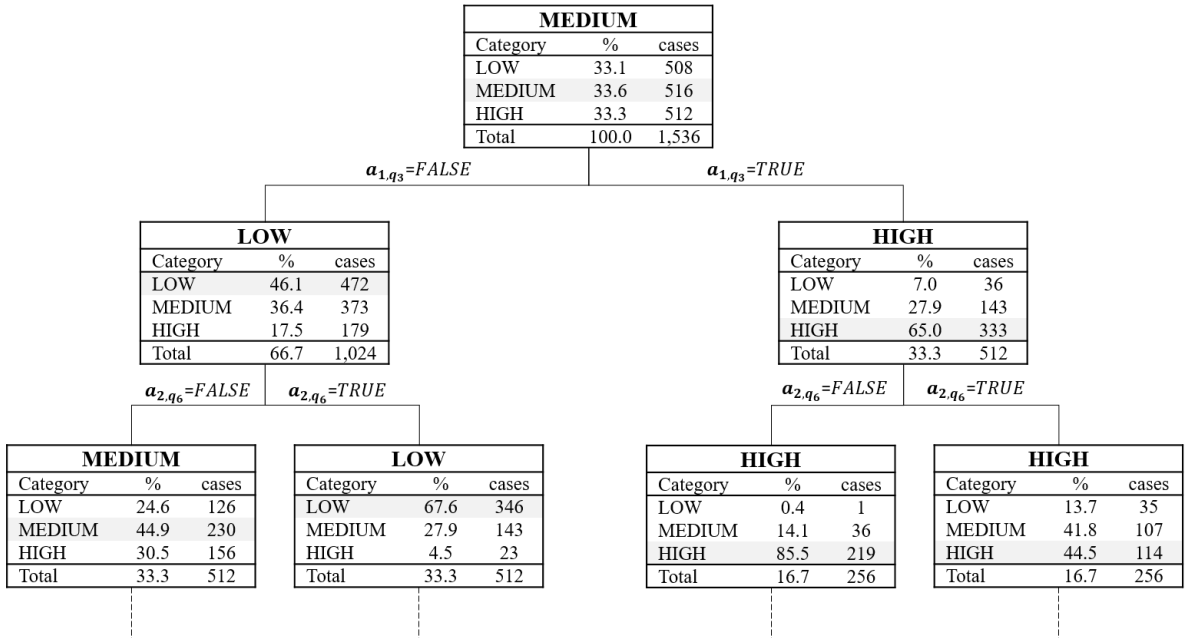

Fig. 3: Pruned decision tree explaining elicited erythema migrans probability model behavior. Each node shows the majority category along with percentage and number of cases belonging to each category. Refer to table 1 for details about the questions and answers. The full tree is available at the link stated in the data statement.

The elicited EM probability scores for all possible cases, detailed decision tree, and FCA context files for all the probability score groups are available at the link stated in the data statement.

The elicited EM probability score can be combined with existing image-only deep learning based EM diagnosis systems to make an effective Lyme disease pre-scanner. Dark skin is underrepresented in existing EM image datasets [7,8]. So, image-only analysis is not appropriate for a proper diagnosis of EM. We believe that combining the elicited probability score from patient data with image-based analysis can partially address this issue.

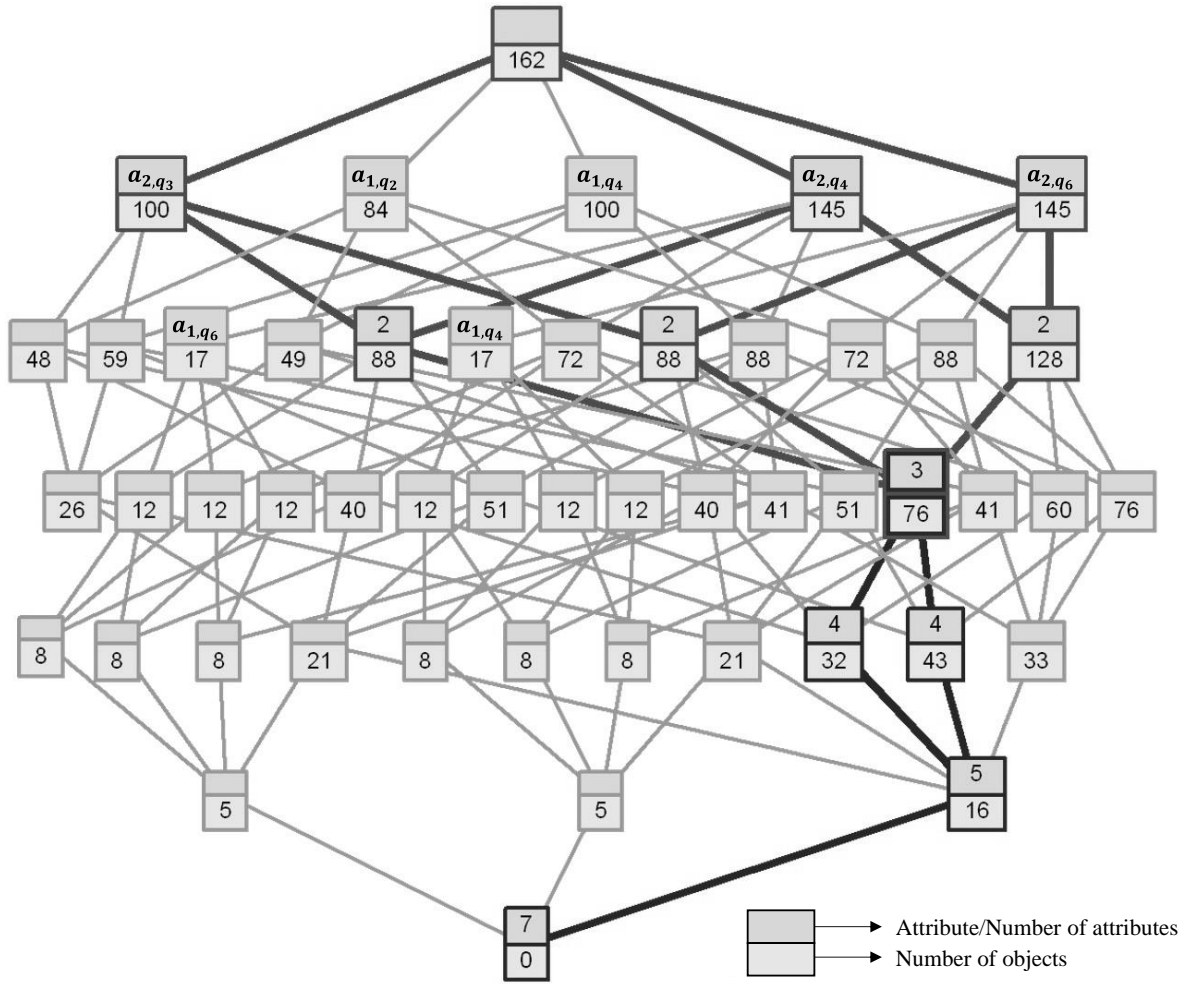

Fig. 4: Formal concept lattice view for 162 very low probability score cases in the range [0,0.1). The top box of a node represents an attribute (answer) or a number of attributes, which are connected by lines, and the bottom box represents how many objects (cases) contain the corresponding attribute shown in the top box. Refer to table 1 for details about the questions and answers.

## 5. Conclusion

In this study, we successfully elicited opinions from fifteen expert doctors to create a model for obtaining EM probability scores from patient data. The elicited probability model will help address the data scarcity problem towards building an effective Lyme disease pre-scanner system. In the future, we will work with the experts on the most appropriate way of combining the elicited EM probability from patient data with probability estimate from deep learning based image classifier to make a robust pre-scanner for Lyme disease.

**Data statement**

The data that support the findings of this study are openly available at https://dappem.limos.fr/elicitation.html.

**Declaration of Competing Interest**

The authors have declared no conflict of interest.

**Acknowledgments**

The authors thank the experts who participated in the expert opinion elicitation process. This work was supported by the European Regional Development Fund, project DAPPEM – AV0021029 (Développement d'une APPlication d'identification des Erythèmes Migrants à partir de photographies). The DAPPEM project coordinated by Olivier Lesens was carried out under the Call for Proposal 'Pack Ambition Research' from the Auvergne-Rhône-Alpes region, France. Mutualité Sociale Agricole (MSA), France also partially funded this research work.